\documentclass[times,review,10pt]{elsarticle}

\usepackage{amssymb}
\usepackage{amsmath}
\usepackage{booktabs} % 用于三线表
\usepackage{multirow} % 用于多行单元格
\usepackage{amsmath}  % 用于数学符号
\usepackage{graphicx}
\usepackage{subcaption}
\usepackage{float}
\usepackage{multirow}
\usepackage{array}
\usepackage{graphicx} % 用于调整表格大小
\usepackage{caption}  % 用于标题控制
\usepackage{makecell} % 用于处理单元格内的换行
\usepackage[table]{xcolor}
\usepackage{adjustbox}
\usepackage{xltabular}
\usepackage{cleveref}
\usepackage{algorithm}
\usepackage{algpseudocode}
\usepackage{makecell}
\usepackage{adjustbox}
\journal{Pattern Recognition}

\begin{document}

\begin{frontmatter}

\title{Vision-Core Guided Contrastive Learning for Balanced Multi-modal Prognosis Prediction of Stroke}

\author{liren Chen, Lidong Sun, Mingyan Huang, Junzhe Tang, Yinghui Zhu, Guanjie Wang,  Yiqing Xia, Ting Xiao*} 

\affiliation{organization={School of Information Science and Engineering, East China University of Science and
 Technology},%Department and Organization
            city={Shanghai},
            postcode={200237}, 
             country={PR China}}
\begin{abstract}
Deep learning and multi-modal fusion have demonstrated transformative potential in medical diagnosis by integrating diverse data sources. However, accurate prognosis for ischemic stroke remains challenging due to limitations in existing multi-modal approaches. First, current methods are predominantly confined to dual-modal fusion, lacking a framework that effectively integrates the trifecta of medical images, structured clinical data, and unstructured text. Second, they often fail to establish deep bidirectional interactions between modalities; To address these critical gaps, this paper proposes a novel tri-modal fusion model for ischemic stroke prognosis. Our approach first enriches the data representation by employing a Large Language Model (LLM) to automatically generate semi-structured diagnostic text from brain MRIs. This process not only addresses the scarcity of expert annotations but also serves as a regularized semantic enhancement, improving multimodal fusion robustness. Furthermore, we design a core component termed the Vision-Conditioned Dual Alignment Fusion Module (VDAFM), which strategically uses visual features as a conditional prior to guide fine-grained interaction with the generated text. This module achieves a dynamic and profound fusion through a dual semantic alignment loss, effectively mitigating modal heterogeneity. Extensive experiments on a real-world clinical dataset demonstrate that our model achieves state-of-the-art performance.
\end{abstract}


%%Research highlights
\begin{highlights}
\item A novel tri-modal framework integrating MRI scans, clinical data, and LLM-generated diagnostic text.
\item The Vision-Conditioned Dual Alignment module enables fine-grained cross-modal fusion.
\item Vision-guided LLM text acts as a semantically enriched and regularized view.
\item Superior prognostic performance achieved through optimized multi-modal interactions.
\end{highlights}

\end{frontmatter}
\section{Introduction}
\label{sec:intro}
Stroke remains a leading global cause of mortality and long-term disability, characterized by high incidence, disability, and mortality rates. Ischemic stroke, which constitutes approximately 65.3\% of all stroke cases \citep{ref1}, represents the predominant subtype. Recent changes in lifestyle patterns combined with accelerated population aging have contributed to a steadily increasing incidence of ischemic stroke \citep{ref2}. The accurate and timely prediction of ischemic stroke prognosis is therefore crucial for developing personalized treatment strategies, optimizing healthcare resource allocation, and ultimately improving patients' quality of life.

Deep learning has been widely applied in the medical field, encompassing areas such as medical image analysis, disease prediction, drug development, and clinical decision-making. In the realm of disease prediction, researchers have developed advanced models to improve predictive accuracy, offering new technical pathways for ischemic stroke prognosis prediction. Traditional approaches to ischemic stroke prognosis primarily rely on single-modal data. For instance, Yu et al. \citep{ref3} extracted and selected 16 key radiomics features from multi-sequence MRI images to develop and compare five machine learning models for predicting patient outcomes. Similarly, Alaka et al. \citep{ref4} employed various machine learning techniques, including random forests \citep{ref5}, decision trees \citep{ref6}, support vector machines \citep{ref7}, and logistic regression \citep{ref8}which were applied to structured clinical data to estimate functional risk after stroke. While these methods have demonstrated reasonable predictive performance, they are inherently limited in their capacity to comprehensively represent the complex clinical status of patients.
Multi-modal fusion technology has emerged as a promising solution that integrates complementary information from diverse data sources. Several recent studies \citep{add1}-\citep{add2} have explored this approach in the context of ischemic stroke prognosis. Ma et al. \citep{ref9} proposed a multi-modal fusion framework named Multitrans, which leverages Transformer architectures and self-attention mechanisms to combine non-contrast computed tomography (NCCT) images with clinical discharge summaries. Brugnara et al. \citep{ref10} integrated clinical, imaging, and angiographic data using a gradient boosting classifier to predict 90-day modified Rankin Scale (mRS) scores, thereby evaluating rehabilitation outcomes following endovascular treatment. Samak et al. \citep{ref11} incorporated 3D NCCT imaging and clinical data into a deep learning model enhanced with channel and spatial attention mechanisms (cSE and sSE), enabling more precise identification of prognostic features.

Although these multi-modal approaches generally outperform single-modal baselines \citep{ref9,ref10,ref11}, they exhibit limitations in both modality selection and fusion strategies \citep{ref12}. First, while medical images, structured clinical data, and unstructured diagnostic texts each provide unique and complementary pathological insights for stroke prognosis, existing research has predominantly been limited to dual-modal frameworks. It is noteworthy that no dedicated model has yet been proposed to integrate all three modalities specifically for stroke prognosis prediction. Furthermore, the unstructured text typically sourced from clinical diagnostic reports is inherently non-standardized and noisy \citep{add4}. Consequently, even with the application of Natural Language Processing (NLP) models like BERT to extract clinical information \citep{ref9}, the generation of high-quality reports continues to rely on expert annotation and is susceptible to inconsistencies in physician documentation. Second, current fusion strategies often underutilize the rich semantic content of medical images.These methods seldom treat visual features as conditional priors to guide the alignment and interaction with textual representations. Specifically, they neglect to project image embeddings into the textual space for joint encoding, a process that could foster deeply interactive cross-modal representations.

To address these limitations, we propose a vision-core guided contrastive learning framework that integrates MRI scans, clinical text, and structured tabular data for balanced multi-modal prognosis prediction of stroke. First, to transcend the semantic limitations of raw pixel data, we deploy a Large Language Model \citep{add5} not merely as a text generator, but as a knowledge injection engine. This module synthesizes a semantically augmented view by projecting implicit MRI features into high-order clinical semantic spaces, thereby incorporating expert priors absent in visual modalities. Crucially, we redefine the inherent variability of LLM generation as a stochastic regularization mechanism. By anchoring these semantic perturbations to visual features, we force the model to capture intrinsic pathological invariants rather than overfitting to deterministic pixel patterns, thus significantly bolstering generalization in data-scarce regimes. Furthermore, we introduce a Vision-Conditioned Dual Alignment Fusion Module (VDAFM) that takes visual features as conditional inputs and performs joint Transformer encoding with textual modalities in the textual space. To preserve informational integrity as much as possible, it further computes a dual cross-semantic loss to contrast the features before and after the joint encoding, thereby achieving enhanced contextual semantics \citep{add6}.

The main contributions of this paper can be summarized as follows:
\begin{itemize}

\item[1.] We propose a tri-modal fusion model that integrates MRI images, clinical text, and tabular data for ischemic stroke prognosis prediction.

\item[2.] We design a Vision-Conditioned Dual Alignment Fusion Module, which projects image features into the textual space and performs joint cross-modal attention to strengthen the interaction between MRI scans and diagnostic reports.

\item[3.] Extensive experiments and ablation studies validate the effectiveness of our approach, demonstrating that tri-modal feature extraction and enhanced image-text fusion significantly improve prognostic performance.

\end{itemize}

\section{Related Work}

\subsection{Single-Modal Prognosis Prediction}
Early research in disease prognosis prediction primarily utilized single modal data. In general medical applications, Liu et al. \citep{ref14} developed a convolutional neural network with multi-view and multi-scale feature fusion for pulmonary nodule classification using CT images. Amit et al. \citep{ref15} employed dedicated convolutional neural networks and pre-trained deep networks to extract features from dynamic contrast-enhanced MRI (DCE-MRI) for breast tumor segmentation and classification. Esteva et al. \citep{ref16} trained deep neural networks on large-scale skin cancer images, achieving classification performance comparable to dermatologists.

In the stroke domain, single-modal approaches follow two main directions. Several studies have focused on imaging features. Yu et al. \citep{ref3} identified 16 radiomics features from multi-sequence MRI and compared multiple classifiers, finding LightGBM to achieve optimal performance. Anne et al. \citep{ref17} developed a deep convolutional network using perfusion-weighted MRI that significantly outperformed conventional models in predicting final infarct volume. Other investigations have centered on clinical variables. Li et al. \citep{ref18} applied multiple machine learning models to structured clinical data to predict 6-month unfavorable outcomes (mRS 3-6), though no model demonstrated clear superiority. In contrast, Heo et al. \citep{ref19} showed that a deep neural network trained on 38 clinical variables significantly outperformed the conventional ASTRAL score in predicting 90-day functional outcomes.

Nevertheless, these methods are all based on single-modal data training, failing to capture the multi-factorial nature of stroke.

In contrast, our method selects representative data from distinct modalities, namely MRI images, clinical diagnostic reports, and data tables, to perform multi-modal data fusion, thereby obtaining more accurate and comprehensive information. 

\subsection{Multi-Modal Fusion for Disease Prognosis}

Multi-modal learning has gained traction in medical prognosis due to its ability to integrate complementary data types. Vale-Silva \citep{ref20} combined clinical, imaging, and multi-omics data within a deep survival model to predict long-term outcomes in pan-cancer patients. Lee et al. \citep{ref21} designed a gated recurrent unit (GRU)-based multi-modal network that processes each data source through separate branches before combining them for predicting conversion from mild cognitive impairment to Alzheimer’s disease. Zarrabi et al. \citep{ref22} derived discriminative features from clinical records and phonocardiogram signals, substantially improving myocardial infarction classification.

In stroke research, multi-modal fusion generally follows one of two paths. The first integrates imaging and unstructured text. Ma et al. \citep{ref9} used Transformer-based encoders (e.g., BERT and ViT) for discharge summaries and NCCT images, fused via self-attention and fully connected networks. The second combines imaging and structured clinical variables. Brugnara et al. \citep{ref10} built a gradient boosting model using clinical, imaging, and angiographic features to predict 90-day mRS. Xiao et al. \citep{ref23} introduced a modal attention mechanism that dynamically weights imaging and clinical features according to their importance, achieving both high accuracy and recall.

However, existing multi-modal studies primarily rely on dimension alignment and feature concatenation \citep{ref24}, leading to the overlook of meaningful inter-modal interactions and the incomplete extraction of clinically relevant lesion information. Our method proposes a vision-guided fusion strategy that projects visual features into the textual space for joint encoding, thereby enabling deeper cross-modal alignment and interaction and facilitating the extraction of more discriminative and clinically relevant representations .

\subsection{Modal Alignment and Fusion Techniques}
Effective multi-modal fusion necessitates specialized alignment and interaction mechanisms. Recent advances in computer vision and natural language processing have introduced several innovative fusion paradigms. The FUSION architecture \citep{ref13} proposed a text-guided visual encoding strategy that injects textual information into the image encoder to achieve finegrained, pixel-level fusion. Wang et al. \citep{ref25} developed SparseMM, which asymmetrically allocates computational resources based on visual relevance, thereby optimizing fusion efficiency in multi-modal large language models. EarthMind \citep{ref26} employed KL divergence loss combined with cross-modal contrastive learning to align image and text representations for segmentation tasks.

In the specific domain of stroke prognosis, researchers have adapted these fusion principles to address clinical challenges. Yoon et al. \citep{ref27} proposed a two-stage fusion strategy where an MM-UNet performs lesion segmentation, after which deep features are fused with hand-crafted radiomics for final classification. Samak et al. \citep{ref11} enhanced feature representation by embedding channel (cSE) and spatial (sSE) attention mechanisms, which dynamically highlight relevant imaging and clinical features. Zihni et al. \citep{add7} designed two feature-level fusion approaches based on unimodal building blocks, in which a 3D CNN was employed to process neuroimaging data and an MLP to handle clinical data. Their findings demonstrated that integrating both data types improves predictive accuracy, while the end-to-end simultaneous training strategy yields more robust performance. 

However, these fusion techniques are primarily developed for natural images and exhibit limitations when applied to medical data \citep{add8}. To address this, we propose a vision-guided text encoding module that leverages the guiding capacity of MRI features to facilitate deeper cross-modal interaction.

\section{Methodology}

\subsection{Overall Framework}

\begin{figure}[!t]%% placement specifier
%% Use \includegraphics command to insert graphic files. Place graphics files in 
%% working directory.
\centering%% For centre alignment of image.
\includegraphics[scale=0.35]{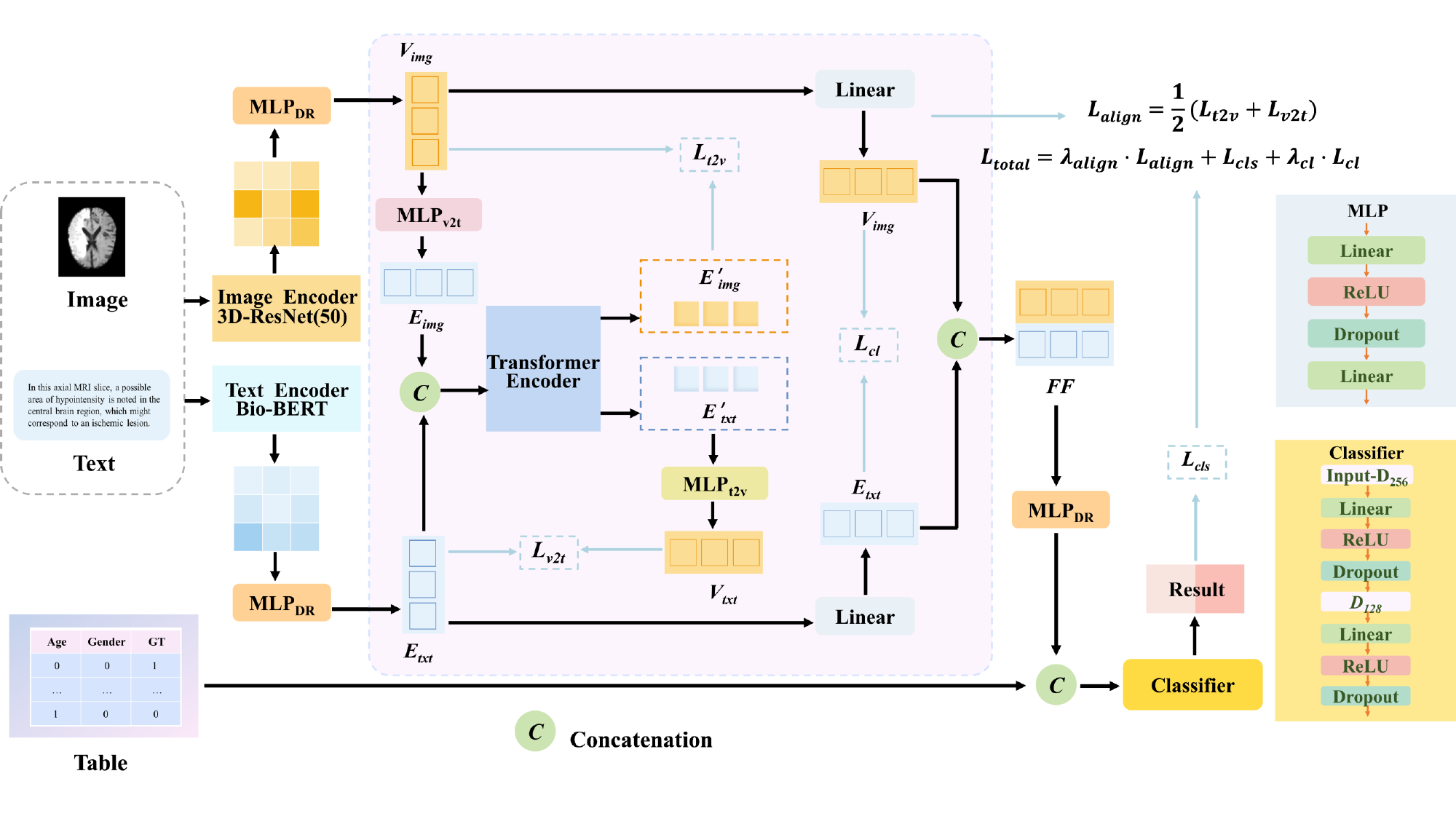}
%% Use \caption command for figure caption and label.
\caption{Cross-modal Deep Fusion Network Architecture for Stroke Prognosis Prediction. }\label{img1}
%% https://en.wikibooks.org/wiki/LaTeX/Importing_Graphics#Importing_external_graphics
\end{figure}

An overview of our model proposed in this paper is illustrated in Figure~ \ref{img1}. This model is a vision-core guided multi-modal fusion model. To achieve differentiated semantic understanding of features from different modalities, the model uses a visual feature extractor (ResNet) and a text feature extractor (Bio-BERT). The input text and image modalities first obtain embedding vectors
through feature extraction. 

To further enhance the fusion between the image and text modalities, we map the dimension-reduced image embedding vectors to a shared text space through a vision-text perceiver, and then concat them with the dimension-reduced text embedding vectors as input for visually-guided text
encoding. To facilitate feature mapping and modal fusion, we propose a dual cross-semantic loss. Guided by this loss, we perform modal fusion between text and image modalities to obtain the fused features. We term this integrated approach, encompassing the vision-guided encoding and the dual-loss guidance, as the Vision-Conditioned Dual Alignment Fusion Module (VDAFM). By introducing VDAFM, the model achieves consistency constraints on multi-modal semantics and efficient fusion of cross-modal information, thereby improving the model’s discriminative ability and interpretability.

At the same time, to reduce the semantic loss of the tabular modality, our framework directly concat the clinical tabular data with the fused feature output by the VDAFM after the fully connected layer, and inputs the concated vector into a classifier to output the classification results.

The methodology of this paper consists of three key components:
\begin{itemize}

\item A visual-anchor-based multimodal semantic enhancement
\item A dual-alignment fusion module for visual conditions
\item A dual semantic cross-constraint mechanism with overall loss optimization

\end{itemize}

\subsection{Multimodal semantic enhancement based on visual anchor points}
The acquisition of high-quality annotated medical datasets remains a formidable challenge, constrained by both high costs and the scarcity of expert clinical knowledge. This data-poor regime has significantly bottlenecked the evolution of medical imaging models. The recent emergence of Large Language Models (LLMs) provides a new paradigm to break this deadlock, where LLM-synthesized MRI diagnostic reports can mitigate the heavy reliance on human-labeled data.

Nevertheless, conventional medical image analysis is typically restricted to the MRI modality, which predominantly captures deterministic local features while overlooking high-level global semantics. Consequently, we introduce a multimodal semantic enhancement framework that utilizes LLM-generated text, anchored by visual features to bridge this gap.

A key contribution of this framework is the redefinition of the LLM-generated text's role: we conceptualize it as a semantically augmented view characterized by stochastic perturbations. By employing the visual image as a "semantic anchor," we ensure textual fidelity while exploiting the vast medical prior knowledge within LLMs to furnish the model with global semantics and clinical descriptors absent in raw imagery. Moreover, we strategically convert the "generative noise" inherent in LLMs into a robust data augmentation tool. This high-order noise serves as a regularization mechanism, bolstering the model’s generalization capabilities across heterogeneous clinical environments.

For practical implementation, we enforce strict constraints to safeguard the anchor's integrity. Five specialized prompts are utilized to mandate that the LLM's inferences—covering infarct core volume, ASPECTS, lesion areas, mass effect, and NIHSS ranges—are strictly grounded in the axial MRI features. Post-generation, we utilize Bio-BERT for sophisticated embedding and implement a "Black and White List" filtering protocol. This ensures that only high-fidelity clinical insights are encoded, while redundant noise and LLM hallucinations are systematically purged. This process can be represented as:

\begin{equation}
Text= \mathrm{LLM}(I, \text{prompt})
\end{equation}
\begin{equation}
X_{\text{txt}} = \mathrm{Bio-BERT}\!\left( 
    \mathrm{Filter}_{\text{black/white}}\!\left( Text \right)
\right)
\end{equation}

\noindent where \textit{Text} denotes the text generated by the large model, \textit{I} represents the input MRI image, and prompt refers to task-specific prompts (e.g., ASPECTS, lesion area, mass effect) provided to the model. LLM stands for the large language model (e.g., qwen-vl-max in this study), $X_\text{txt}$ is the text embedding, Bio-BERT refers to the Bio-BERT encoder, and Whitelist pertains to the whitelist/blacklist processing.

This black-and-white list filtering protocol effectively purges invalid generative noise, thereby maximizing the SNR and fidelity of the encoded embeddings. In parallel, our randomized prompt sampling strategy bolsters the regularization mechanism: by injecting controllable semantic perturbations, it prevents the rote memorization of text patterns and forces the model to capture intrinsic pathological features. Consequently, we treat the text as a distinct high-order semantic modality. Through cross-modal fusion with MRI data, this approach bridges the semantic gap inherent in unimodal visual analysis, equipping the model with superior deep reasoning capabilities and interpretability for real-world clinical applications.

\subsection{Vision-Conditioned Dual Alignment Fusion Module}
\subsubsection{Dimensionality Alignment}
To capture feature information from different semantic spaces, in addition to the text feature extractor (Bio-BERT) mentioned above, we also use a dedicated visual feature extractor (ResNet) to convert the input image modality into an embedding vector and use MLP mapping for dimensionality alignment. The preprocessed brain MRI image $I$ of a stroke patient is
passed through a ResNet encoder to extract the visual embedding:

\begin{equation}
X_{\text{img}} = \mathrm{ResNet}(I)
\end{equation}

Since the output image embedding vector and the text embedding vector differ greatly in dimensionality, we apply MLP-based mappings to reduce the dimensions of both modalities for more convenient subsequent processing. The resulting dimension-reduced visual embedding and diagnostic-report text embedding are expressed by the following equations:
\begin{equation}
V_{\text{img}} = \mathrm{MLP}_{\text{DR}}\!\left( X_{\text{img}} \right)
\end{equation}

\begin{equation}
E_{\text{txt}} = \mathrm{MLP}_{\text{DR}}\!\left( X_{\text{txt}} \right)
\end{equation}

\noindent where $\mathrm{MLP_{\mathrm{DR}}}(\cdot)$ represents the operation of Dimensionality Reduction (DR) performed by a Multi-Layer Perceptron (MLP).

\subsubsection{Visually-Guided Text Encoding}
Next, unlike traditional vision-language modal fusion methods that directly concat after dimensionality reduction, We introduce a visual-guided text alignment module to achieve cross-modal alignment between text features and image features through spatial interaction, in order to address the inherent heterogeneity between continuous visual manifolds and discrete text semantic spaces:

\begin{equation}
E_{\text{img}} = \mathrm{MLP}_{\text{v2i}}\!\left( V_{\text{img}} \right)
\end{equation}

\noindent where $\mathrm{MLP_{\mathrm{v2i}}(\cdot)}$ refers to a multi-layer perceptron (MLP) that maps visual feature vectors to the text feature space.

For subsequent Transformer encoding, we concat it with the text
embedding in the text feature space to obtain the concated vector:

\begin{equation}
E_{\text{img-txt}} = {C}\!\left( E_{\text{img}},\, E_{\text{txt}} \right)
\end{equation}
\noindent where \(C\) denotes the concat operation.

To further extract important information related to the image and text,
we use the visual embedding as a guide and jointly encode them through a
transformer encoding layer:

\begin{equation}
\left( E'_{\text{img}},\, E'_{\text{txt}} \right)
    = f_{\text{Trans}}\!\left( E_{\text{img}},\, E_{\text{txt}} \right)
\end{equation}

\noindent where $E'_{\text{img}}$ and $E'_{\text{txt}}$ represent the visual and text embeddings after Transformer encoding, and  $f_{\text{Trans}}(\cdot)$represents the Transformer encoding.
\subsubsection{Dual Cross-Semantic Loss}
To alleviate the modality discrepancy during feature mapping and the modal difference during modal fusion, we interactively map the visual and text features to calculate a dual cross-loss. It is essential to ensure that the visual features $E'_{img}$ produced by the transformer encoder remain highly consistent with the embedding vectors $V_{img}$ in the visual feature space. To mitigate semantic loss during encoding, we compute a vision–text space mapping contrastive loss between the encoded visual vector and the corresponding image embedding vector in the visual feature space:

\begin{equation}
L_{\text{t2v}} 
= 1 - 
\frac{
    E'_{\text{img}} \cdot V_{\text{img}}
}{
    \left\lVert E'_{\text{img}} \right\rVert_{2}
    \left\lVert V_{\text{img}} \right\rVert_{2}
}
\end{equation}

To ease subsequent loss calculation and  keep the loss weights consis-
tent, we also apply an MLP mapping operation to the text embedding vector
after encoding to obtain the text vector mapped to the visual dimension:

\begin{equation}
V_{\text{txt}} = \mathrm{MLP}_{\text{t2v}}\!\left( E'_{\text{txt}} \right)
\end{equation}

Similarly, the vector $V_{txt}$ after mapping should be highly similar to the text representation in the text feature space $E_{txt}$. We calculate the text-vision space contrastive loss between $V_{txt}$ and $E_{txt}$:
\begin{equation}
L_{\text{v2t}} 
= 1 -
\frac{
    E_{\text{txt}} \cdot V_{\text{txt}}
}{
    \left\lVert E_{\text{txt}} \right\rVert_{2}\,
    \left\lVert V_{\text{txt}} \right\rVert_{2}
}
\end{equation}

Considering both semantic losses, the total semantic loss $L_{align}$ is:
\begin{equation}
L_{\text{align}}
= \frac{1}{2}\left( L_{\text{t2v}} + L_{\text{v2t}} \right)
\end{equation}

By mutually constraining and aligning the text and visual modalities,this dual cross-semantic loss can reduce semantic loss during feature extraction and multi-modal fusion, improving the model's ability to understand and integrate multi-modal information.

\subsubsection{Multimodal Fusion}
To comprehensively consider the text and image modalities for the final classification, we concat them after a linear layer to obtain the context feature vector
after visually-guided modal fusion:
\begin{equation}
F_{\text{f}}
= C\!\left( L(V_{\text{img}}),\, L(E_{\text{txt}}) \right)
\end{equation}

\noindent where $F_{\text{f}}$ represents the fused feature vector and $L(\cdot)$ represents processing through a fully connected layer.
The fused input is fed into a classifier for classification to obtain the final prognosis prediction result. For effective supervised classification, we calculate the classification loss:
\begin{equation}
L_{\text{cls}}
= -\bigl[\, y \log p + (1 - y)\log(1 - p) \,\bigr]
\end{equation}

\noindent where \textit{y} is the true class and \textit{p} is the predicted class.

\subsubsection{Contrastive Loss}
To better achieve modal fusion, we further align the image and text modalities in a shared semantic space to achieve cross-modal semantic consistency. We calculate the contrastive loss:
\begin{equation}
\begin{aligned}
L_{\text{cl}}
&= \frac{1}{2}\left( L_{\text{img-txt}} + L_{\text{txt-img}} \right) \\
&= -\frac{1}{2N}
\Biggl[
\sum_{i=1}^{N}
\log\frac{\exp(S_{ii})}{\sum_{j=1}^{N}\exp(S_{ij})}
+
\sum_{j=1}^{N}
\log\frac{\exp(S_{jj})}{\sum_{i=1}^{N}\exp(S_{ij})}
\Biggr]
\end{aligned}
\end{equation}

\noindent where \textit{N} is the number of image-text pairs, $L_{\text{img-txt}}$ is the loss for image-to-text matching, and $L_{\text{txt-img}}$ is the loss for text-to-image matching. $S_{ij} = \frac{\tilde{v}_i^T \tilde{e}_j}{\tau} \quad (\tau > 0)$ is the similarity between the image and text, $\tilde{v}_i^T$ is the embedding vector of the \textit{i}-th image, and $\tilde{e}_i$ is the embedding vector of the \textit{j}-th text. \textit{j} denotes the embedding vector of the \textit{j}-th text, $\tau$ is the temperature parameter.

\subsection{Overall Loss}
The overall loss of our multi-modal fusion model is:
\begin{equation}
L_{\text{total}}
= \lambda_{\text{align}} L_{\text{align}}
+ L_{\text{cls}}
+ \lambda_{\text{cl}} L_{\text{cl}}
\end{equation}

\noindent where $\lambda_{align}$ is the weight coefficient for the alignment loss, $L_{align}$ is the semantic loss,it can reduce the semantic loss during Transformer encoding and guide the model for accurate classification. $L_{cls}$ is the classification loss,which can guide the module for accurate classification. $\lambda_{cl}$ is the weight coefficient for the contrastive learning loss, and $L_{cl}$ is the contrastive learning loss, it is used to further strengthen the semantic alignment between image and text features before fusion. By introducing multiple loss terms, the model can achieve a balance between semantic preservation, modal alignment, and accurate classification. This combination of multi-task losses allows the model to optimize multiple objectives simultaneously, ultimately improving the overall performance of multi-modal tasks.

\section{Experiments and Results}

\subsection{Preprocessing}
\subsubsection{Data Collection and Preprocessing}
The multi-modal data used in this study (including MRI images and clinical tabular data) were all from Shanghai Tongji Hospital and Xinhua Hospital. After removing samples where image and tabular information did not correspond, a total of 729 valid cases were included. The clinical data information is stored in tables and includes patient age, gender, and relevant clinical information about the past medical history. 20\% of the entire dataset (146 cases) was used as the test set, and the remaining 80\% were used as training data. Within the training data, 20\% (117 cases) was selected as the validation set, and the rest was the training set (466 cases).
For missing values in the tabular modality, the mode imputation method was used for preprocessing to ensure data integrity. At the same time, according to clinical guidelines, we labeled cases with an NIHSS score < 7 as having a good prognosis (label = 1) and cases with an NIHSS score $\ge$ 7 as having a poor prognosis (label = 0), thus constructing a binary classification label for subsequent prognosis prediction modeling.
For the image modality, we conducted strict quality control screening, removing 3D MRI images with artifacts or low quality, and retaining only high-quality images for subsequent analysis.
The text modality data was automatically generated as diagnostic reports using qwen-vl-max with prompt generation. We used self-designed prompts. The model generation parameters were set to max\_tokens  250, top\_p  0.8 and temperature 0.7, in order to balance the professionalism and diversity of the generated text.
In the feature extraction stage, the image data was encoded using the Tencent Medical pre-trained model MedicalNet (based on the ResNet-50 structure). The text reports were processed using the BioBERT(v1.1) model to extract semantic features. We used self-designed prompts and selected the last four layers corresponding to the Classification Special Token layers of the model for concatenation as the encoding result. During text encoding, a strategy of randomly selecting prompts was adopted to increase the diversity and robustness of the text encoding.

\subsubsection{Data Augmentation}
To address the class imbalance problem in the dataset (label=0 accounts for approximately 68.14\% of the total), we adopted the following data augmentation strategies. For the tabular modality, we used the nearest neighbor replication SMOTE strategy (using KNN search for the nearest, K=1). This ensured the alignment of the augmented data modalities. During training, Gaussian noise with an intensity of 0.01 was injected into the input features with a probability of 50\%. At the same time, MixUp was applied with the same probability, with its mixing coefficient $\lambda$ following a Beta(0.4, 0.4) distribution.

\subsection{Experimental Setup}
In our experiments, the model was built using the PyTorch framework and trained on a single NVIDIA GeForce RTX 4090 GPU. We used the RAdam optimizer  with an initial learning rate of 1e-4 and a weight decay of 0. A gradient clipping magnitude of 1.0 was applied to prevent gradient explosion. The model was trained for a maximum of 100 epochs, utilizing an early stopping mechanism to prevent overfitting. Training was terminated if the validation set AUC failed to improve for 30 consecutive epochs. In the loss function, the alignment loss coefficient was set to $\lambda_{align}$=0.02, and the contrastive loss coefficient was set to $\lambda_{cl}$=0.2, with the latter decaying at an exponential rate of 0.95 during training. The learning rate schedule employed a cosine annealing strategy with a linear warmup for the first 15 epochs. Furthermore, the optimal hyperparameters were determined via Grid Search, and the model's final performance was rigorously evaluated using 5-fold cross-validation. To account for stochasticity, all experiments were repeated 4 times with random seeds from 39 to 42, and the final average results and standard deviations are reported. The model's performance was comprehensively evaluated using several common metrics for medical binary classification to fully assess its classification ability. These included area under the curve (AUC), accuracy(ACC), F1-Score(F1), precision, recall, and specificity.

\subsection{Experimental Results and Analysis}
Since there is currently no tri-modal model specifically for predicting stroke prognosis, we used the following three comparison methods:
(1). Comparison with conventional  single-modal prediction baseline, including traditional machine learning algorithms such as XGBoost\citep{ref28}, Random Forest\citep{ref5}, and Support Vector Machine\citep{ref7}.
(2). Comparison with deep learning methods. We input the tabular modality information and the extracted image information into a unified classifier and built a simple neural network for comparison, such as MLP\citep{ref29}, TableNet, and ImageNet.
(3). Comparison with existing multi-modal methods. We independently built multi-modal networks with simple fusion mechanisms like Concat and Cross-attention as comparison baselines. Given the current lack of dedicated tri-modal models for stroke prognosis prediction, we performed adaptive modifications on three representative dual-modal frameworks: EarthMind\citep{ref26}, Fusion\citep{ref13}, and the dual-modal optimization framework SparseMM\citep{ref25}.Our unified adaptation strategy involved two main steps: First, we extended their input mechanisms to accommodate our three modalities (Image, Text, and Tabular), specifically by concatenating the fused Image-Text features with the Tabular features. Second, we replaced or removed the original downstream task heads of these models (e.g., the generative head of SparseMM or other model-specific heads), uniformly adding a binary classifier to suit our discriminative prediction task. Through this approach, we transformed these existing dual-modal frameworks into comparative models capable of tri-modal classification prediction. 

\begin{table}[!t]
\centering
\caption{Comprehensive Performance Comparison with Baseline Methods}
\label{tab1}
\large
\renewcommand{\arraystretch}{1.8}
\setlength{\tabcolsep}{6pt}
\begin{adjustbox}{max width=\textwidth}
\begin{tabular}{@{}l l c c c c c c@{}}
\hline
\textbf{Category} & \textbf{Method} & \textbf{AUC} & \textbf{ACC} & \textbf{F1} & \textbf{Recall} & \textbf{Precision} & \textbf{Specificity} \\
\hline
\multirow{3}{*}{\makecell[l]{Traditional\\Methods}} 
& XGBoost & $71.07 \pm 4.39$ & $73.48 \pm 0.88$ & $42.82 \pm 2.81$ & $31.38 \pm 3.15$ & $68.15 \pm 2.81$ & $93.07 \pm 1.57$ \\
& Random Forest (RF) & $71.46 \pm 5.69$ & $73.14 \pm 4.04$ & $53.80 \pm 5.62$ & $48.94 \pm 3.98$ & $59.96 \pm 8.44$ & $84.41 \pm 4.39$ \\
& SVM & $70.28 \pm 1.93$ & $70.61 \pm 2.72$ & $52.34 \pm 5.54$ & $51.06 \pm 6.38$ & $53.74 \pm 4.55$ & $79.70 \pm 1.11$ \\
\hline
\multirow{3}{*}{\makecell[l]{Deep\\Learning\\Methods}} 
& MLP & $71.08 \pm 3.18$ & $73.99 \pm 1.01$ & $45.56 \pm 4.51$ & $34.57 \pm 5.07$ & $67.79 \pm 2.66$ & $92.33 \pm 1.46$ \\
& TableNet & $72.03 \pm 4.20$ & $74.83 \pm 2.10$ & $52.02 \pm 4.74$ & $43.09 \pm 4.61$ & $65.71 \pm 4.64$ & $89.60 \pm 1.11$ \\
& Imagenet & $73.08 \pm 1.04$ & $73.80 \pm 0.89$ & $50.12 \pm 4.23$ & $42.39 \pm 6.96$ & $63.72 \pm 5.98$ & $88.25 \pm 3.96$ \\
\hline
\multirow{5}{*}{\makecell[l]{Multi-modal\\Strategies}} 
& Concat & $78.46 \pm 2.82$ & $77.40 \pm 1.94$ & $64.89 \pm 2.34$ & $66.30 \pm 4.48$ & $63.92 \pm 3.96$ & $82.50 \pm 3.77$ \\
& Attention & $79.86 \pm 1.00$ & $74.32 \pm 4.90$ & $62.41 \pm 2.80$ & $66.85 \pm 6.22$ & $60.31 \pm 9.78$ & $77.75 \pm 9.73$ \\
& Earthmind & $79.34 \pm 1.82$ & $77.74 \pm 3.01$ & $65.25 \pm 4.46$ & $66.30 \pm 4.98$ & $64.37 \pm 4.92$ & $83.00 \pm 3.08$ \\
& Fusion & $79.39 \pm 2.89$ & $78.08 \pm 4.14$ & $67.09 \pm 4.36$ & $70.11 \pm 0.94$ & $64.70 \pm 7.58$ & $81.75 \pm 5.80$ \\
& Sparsemm & $79.94 \pm 1.86$ & $78.72 \pm 2.84$ & $66.22 \pm 3.62$ & $66.85 \pm 2.37$ & $65.71 \pm 5.29$ & $83.75 \pm 3.34$ \\
\hline
\makecell[l]{Proposed\\Method} & Ours & $\mathbf{81.06 \pm 1.97}$ & $\mathbf{81.16 \pm 2.30}$ & $\mathbf{69.26 \pm 3.68}$ & $\mathbf{67.39 \pm 4.35}$ & $\mathbf{71.39 \pm 4.19}$ & $\mathbf{87.50 \pm 2.29}$ \\
\hline
\end{tabular}
\end{adjustbox}
\end{table}
As shown in Table \ref{tab1}, we systematically compared our model with baseline methods across different levels:

In comparison with Traditional Machine Learning Baselines, although XGBoost achieved the highest specificity, its recall was the lowest, indicating that the model adopted an overly conservative classification strategy that sacrifices sensitivity to maintain negative sample accuracy. In contrast, by effectively fusing image, text, and clinical data, our model achieved an AUC of 81.06\% and an ACC of 81.16\%, surpassing traditional baselines (e.g., XGBoost, Random Forest, and SVM) by approximately 10 percentage points. Notably, our model attained an F1-score of 69.26\%, an improvement of 15 to 26 percentage points over traditional methods. This demonstrates a superior capability in capturing true positive cases and reducing the risk of misclassifying high-risk patients as low-risk, aligning well with the critical clinical requirement of "minimizing missed diagnoses.

Comparison with Single-modal Deep Learning Methods We observed that the single-modal MLP model, lacking an effective fusion mechanism, replicated the conservative limitations seen in XGBoost. Despite a high specificity of 92.33\%, its recall was only 34.57\%, resulting in an F1-score lower than all single-modal baselines. Conversely, benefiting from cross-modal information interaction, our model not only maintained leads in AUC and ACC but also achieved a recall of 67.39\%—an increase of over 24 percentage points compared to the best single-modal baseline—and led in F1-score by approximately 17 percentage points. These results strongly confirm that our multi-modal fusion mechanism significantly outperforms deep learning models relying on single data sources, particularly in capturing key samples and balancing overall classification performance.

Comparison with Current Multi-modal Fusion Methods Compared to simple feature concatenation (Concat), basic attention mechanisms, and mainstream methods such as EarthMind, Fusion, and SparseMM, our model demonstrated superior robustness. Specifically, against the second-best performing method, SparseMM (AUC 79.94\%), our model improved the AUC by 1.12 percentage points. Furthermore, the F1-score (69.26\%) and precision (71.39\%) increased by 3.04 and 5.68 percentage points, respectively, highlighting the comprehensive advantage of our image-text feature fusion. Regarding the FUSION method, although it achieved the highest recall (70.11\%), its relatively low precision (64.70\%) suggests a higher rate of false positives. In contrast, our model achieved optimal precision (71.39\%) and specificity (87.50\%) while maintaining a competitive recall (67.39\%). This indicates that by employing a visual-centric fusion strategy combined with contrastive learning and dynamic weight adjustment, our model effectively controls false positives and enhances feature discrimination, thereby achieving the best trade-off between recall and precision.

Beyond these external performance metrics, we further conducted an internal representation analysis using the test set (fixed with random seed 42) to evaluate the model's alignment efficacy. Our results demonstrate that after VDAFM-guided transformer joint encoding, the cross-modal cosine similarity angle consistently decreased by $9.6^\circ$, compared to non-aligned baselines. This quantitative evidence confirms that our framework successfully bridges the modality gap, fostering a more compact and consistent shared semantic space. Such enhanced internal alignment directly underpins the model’s superior discriminative stability and its ability to effectively handle complex clinical scenarios by synergizing visual anchors with high-level textual representations.
\subsection{Ablation Study of Various Components}
To verify the effectiveness of each modality's information in multi-modal information fusion, we designed the following modal ablation experiments: single-modal comparison, which included single-modal experiments for table, image, and text; dual-modal comparison, which included dual-modal experiments for table and image, table and text, and image and text. The specific ablation method was to modify the dataloader for the corresponding ablated modality so that it returned a zero matrix. At the same time, to verify the effectiveness of the loss module components, we conducted relevant ablation experiments on the loss, disabling the semantic alignment loss and the contrastive loss, respectively. The experimental results are presented in Table \ref{tab2}. 

\begin{table}[!t]
\centering
\caption{Ablation Study on Modality Combinations}
\begin{adjustbox}{max width=\textwidth}
\renewcommand{\arraystretch}{1.3}
\begin{tabular}{ccc|cccccc}
\hline
\multicolumn{3}{c|}{Modality} & \multicolumn{6}{c}{Evaluation Metrics} \\
\cline{1-3} \cline{4-9}
Text & Image & Table & AUC & ACC & F1 & Recall & Precision & Specificity \\
\hline
\multirow{2}{*}{$\checkmark$} & \multirow{2}{*}{$\times$} & \multirow{2}{*}{$\times$} & 49.51 & 38.87 & 45.85 & 83.15 & 31.85 & 18.50 \\
 & & & {\footnotesize($\pm$0.92)} & {\footnotesize($\pm$3.40)} & {\footnotesize($\pm$3.13)} & {\footnotesize($\pm$13.26)} & {\footnotesize($\pm$0.97)} & {\footnotesize($\pm$10.69)} \\[2pt]
\multirow{2}{*}{$\times$} & \multirow{2}{*}{$\checkmark$} & \multirow{2}{*}{$\times$} & 67.12 & 66.27 & 55.33 & 67.93 & 50.05 & 65.50 \\
 & & & {\footnotesize($\pm$3.08)} & {\footnotesize($\pm$4.75)} & {\footnotesize($\pm$3.12)} & {\footnotesize($\pm$16.15)} & {\footnotesize($\pm$7.75)} & {\footnotesize($\pm$14.36)} \\[2pt]
\multirow{2}{*}{$\times$} & \multirow{2}{*}{$\times$} & \multirow{2}{*}{$\checkmark$} & 77.58 & 75.00 & 61.92 & 64.67 & 60.51 & 79.75 \\
 & & & {\footnotesize($\pm$3.93)} & {\footnotesize($\pm$3.77)} & {\footnotesize($\pm$4.67)} & {\footnotesize($\pm$8.75)} & {\footnotesize($\pm$6.90)} & {\footnotesize($\pm$6.76)} \\[2pt]
\multirow{2}{*}{$\checkmark$} & \multirow{2}{*}{$\checkmark$} & \multirow{2}{*}{$\times$} & 73.89 & 69.01 & 56.60 & 64.13 & 51.64 & 71.25 \\
 & & & {\footnotesize($\pm$2.01)} & {\footnotesize($\pm$3.67)} & {\footnotesize($\pm$2.11)} & {\footnotesize($\pm$7.29)} & {\footnotesize($\pm$5.68)} & {\footnotesize($\pm$8.04)} \\[2pt]
\multirow{2}{*}{$\checkmark$} & \multirow{2}{*}{$\times$} & \multirow{2}{*}{$\checkmark$} & 79.40 & 76.88 & 65.20 & 67.93 & 63.20 & 81.00 \\
 & & & {\footnotesize($\pm$1.70)} & {\footnotesize($\pm$4.92)} & {\footnotesize($\pm$5.17)} & {\footnotesize($\pm$4.17)} & {\footnotesize($\pm$7.80)} & {\footnotesize($\pm$7.04)} \\[2pt]
\multirow{2}{*}{$\times$} & \multirow{2}{*}{$\checkmark$} & \multirow{2}{*}{$\checkmark$} & 79.97 & 76.37 & 63.21 & 64.13 & 62.77 & 82.00 \\
 & & & {\footnotesize($\pm$1.35)} & {\footnotesize($\pm$2.67)} & {\footnotesize($\pm$1.76)} & {\footnotesize($\pm$2.43)} & {\footnotesize($\pm$5.26)} & {\footnotesize($\pm$4.95)} \\[2pt]
\multirow{2}{*}{$\checkmark$} & \multirow{2}{*}{$\checkmark$} & \multirow{2}{*}{$\checkmark$} & \textbf{81.06} & \textbf{81.16} & \textbf{69.26} & \textbf{67.39} & \textbf{71.39} & \textbf{87.50} \\
 & & & {\footnotesize\textbf{($\pm$1.97)}} & {\footnotesize\textbf{($\pm$2.30)}} & {\footnotesize\textbf{($\pm$3.68)}} & {\footnotesize\textbf{($\pm$4.35)}} & {\footnotesize\textbf{($\pm$4.19)}} & {\footnotesize\textbf{($\pm$2.29)}} \\
\hline
\end{tabular}
\end{adjustbox}
\label{tab2}
\end{table}

\subsubsection{Single-modality Ablation}
In the single-modality comparison, the table modality (Table) showed the strongest predictive ability, with its AUC (77.58\%) and ACC (75.00\%) significantly better than the image and text modalities, indicating that structured clinical data plays a fundamental role in this task. However, the performance of all single modalities was far lower than that of the complete multi-modal model (Ours), reflecting the necessity of multi-modal fusion. It is particularly noteworthy that although the text modality (Text) obtained a high recall (83.15\%), its extremely low precision (31.85\%) and specificity (18.50\%) reflect that the model adopted a "majority positive" conservative strategy due to limited feature discrimination, leading to a large number of false positives. In contrast, the fusion model proposed in this paper, by introducing stable and discriminative features from the table and image modalities, significantly improved the accuracy and robustness of classification while maintaining a high recall, ultimately achieving comprehensive optimality in all indicators. This not only proves that multi-modal fusion can effectively integrate the advantages of different data sources, but also reveals that its key lies in achieving information complementarity between modalities, thereby breaking through the performance ceiling of a single perspective and obtaining more robust and accurate judgments.

\begin{table}[!t]
\centering
\caption{Ablation Study on Loss Function}
\begin{adjustbox}{max width=\textwidth}
\renewcommand{\arraystretch}{1.3}
\begin{tabular}{ccc|cccccc}
\hline
\multicolumn{3}{c|}{Loss Function} & \multicolumn{6}{c}{Evaluation Metrics} \\
\cline{1-3} \cline{4-9}
\(L_{clr}\) & \(L_{cl}\) & \(L_{align}\) & AUC & ACC & F1 & Recall & Precision & Specificity \\
\hline
\multirow{2}{*}{$\checkmark$} & \multirow{2}{*}{$\times$} & \multirow{2}{*}{$\times$} & 79.91 & 78.77 & 67.26 & 69.02 & 65.72 & 83.25 \\
 & & & {\footnotesize($\pm$1.46)} & {\footnotesize($\pm$2.70)} & {\footnotesize($\pm$3.37)} & {\footnotesize($\pm$2.82)} & {\footnotesize($\pm$4.73)} & {\footnotesize($\pm$3.34)} \\[2pt]
\multirow{2}{*}{$\checkmark$} & \multirow{2}{*}{$\checkmark$} & \multirow{2}{*}{$\times$} & 80.09 & 79.11 & 67.44 & 68.48 & 66.53 & 84.00 \\
 & & & {\footnotesize($\pm$1.60)} & {\footnotesize($\pm$2.49)} & {\footnotesize($\pm$3.17)} & {\footnotesize($\pm$2.43)} & {\footnotesize($\pm$4.40)} & {\footnotesize($\pm$3.00)} \\[2pt]
\multirow{2}{*}{$\checkmark$} & \multirow{2}{*}{$\times$} & \multirow{2}{*}{$\checkmark$} & 80.38 & 79.62 & 67.99 & 68.48 & 67.70 & 84.75 \\
 & & & {\footnotesize($\pm$1.84)} & {\footnotesize($\pm$2.24)} & {\footnotesize($\pm$2.24)} & {\footnotesize($\pm$1.88)} & {\footnotesize($\pm$4.39)} & {\footnotesize($\pm$3.34)} \\[2pt]
\multirow{2}{*}{$\checkmark$} & \multirow{2}{*}{$\checkmark$} & \multirow{2}{*}{$\checkmark$} & \textbf{81.06} & \textbf{81.16} & \textbf{69.26} & \textbf{67.39} & \textbf{71.39} & \textbf{87.50} \\
 & & & {\footnotesize\textbf{($\pm$1.97)}} & {\footnotesize\textbf{($\pm$2.30)}} & {\footnotesize\textbf{($\pm$3.68)}} & {\footnotesize\textbf{($\pm$4.35)}} & {\footnotesize\textbf{($\pm$4.19)}} & {\footnotesize\textbf{($\pm$2.29)}} \\
\hline
\end{tabular}
\end{adjustbox}
\label{tab3}
\end{table}

\subsubsection{Dual-modality Ablation}
We examined the effect of dual-modality combinations. The performance of all dual-modality combinations (e.g., Table+Image) was better than the best single-modality, which verifies the effectiveness of information complementarity between modalities. Among them, the combinations containing tabular data (Table+Image, Text+Table) performed best, but their AUC (79.97\%) and F1-score (65.20\%) were still lower than the OURS model (AUC: 81.06\%, F1: 69.26\%). It is worth noting that OURS achieved the greatest advantage in precision (71.39\%) and specificity (87.50\%), which indicates that our fusion method can more effectively suppress false positives and improve the discriminative reliability of the model.

\begin{figure}[!t]
\centering
    \begin{subfigure}{0.99\textwidth}
        \includegraphics[width=1\linewidth]{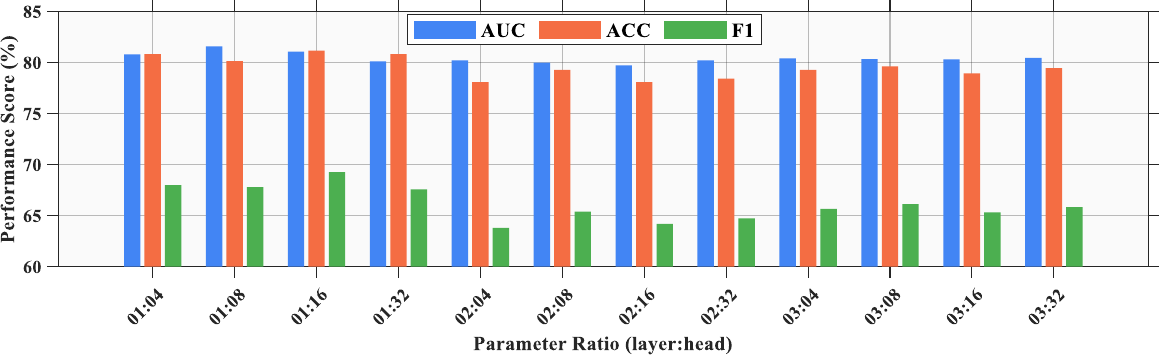}
     \end{subfigure}

     \begin{subfigure}{0.99\textwidth}
        \includegraphics[width=1\linewidth]{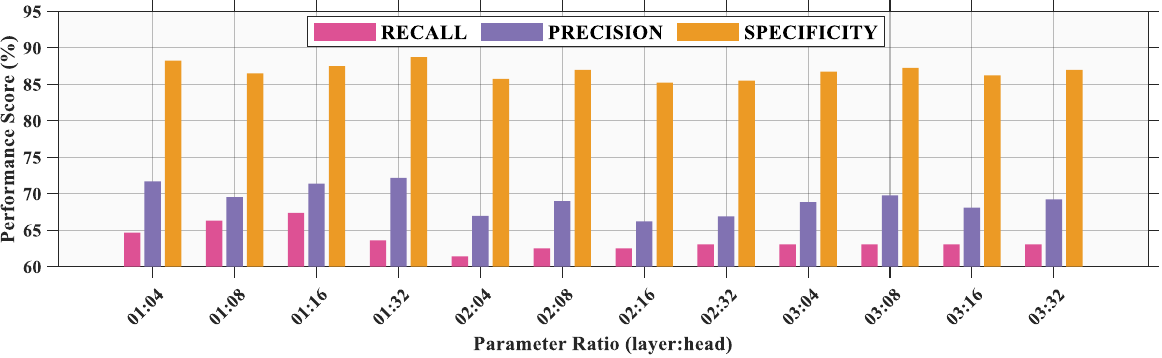}
     \end{subfigure}
     
      \caption{Sensitivity analysis results of encoder parameters}
      \label{img2}
     
\end{figure}

\subsubsection{Loss Ablation}
In the loss function ablation study, each component showed clear value. As shown in Table \ref{tab3}. When only the classification loss (\(L_{clr}\)) was used, the model obtained the highest recall (69.02\%), which indicates that in the absence of additional constraints, the model tends to adopt a looser decision boundary. Although it can capture more positive examples, it also leads to lower precision and specificity. After gradually introducing the contrastive loss (\(L_{cl}\)) and the alignment loss (\(L_{align}\)), the model gradually optimized indicators such as AUC, precision, and specificity by enhancing feature discrimination and modal consistency. Finally, the complete loss function achieved the best results on all key indicators, demonstrating a comprehensive improvement in overall performance. Notably, it yielded significant improvements in precision (71.39\%) and specificity (87.50\%), indicating that it effectively balanced recognition sensitivity and classification.

\subsection{Hyperparameter Study of the VDAFM}

\begin{figure}[!t]
\centering
    \begin{subfigure}{0.99\textwidth}
        \includegraphics[width=1\linewidth]{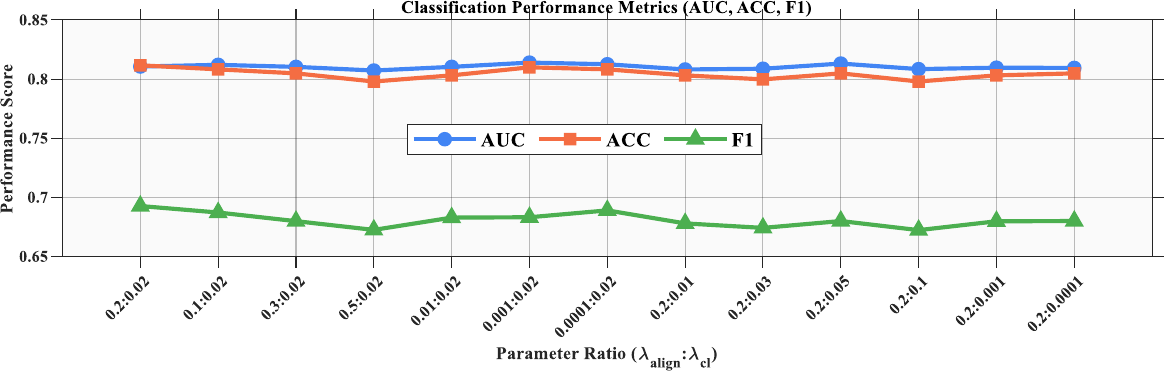}
     \end{subfigure}

     \begin{subfigure}{0.99\textwidth}
        \includegraphics[width=1\linewidth]{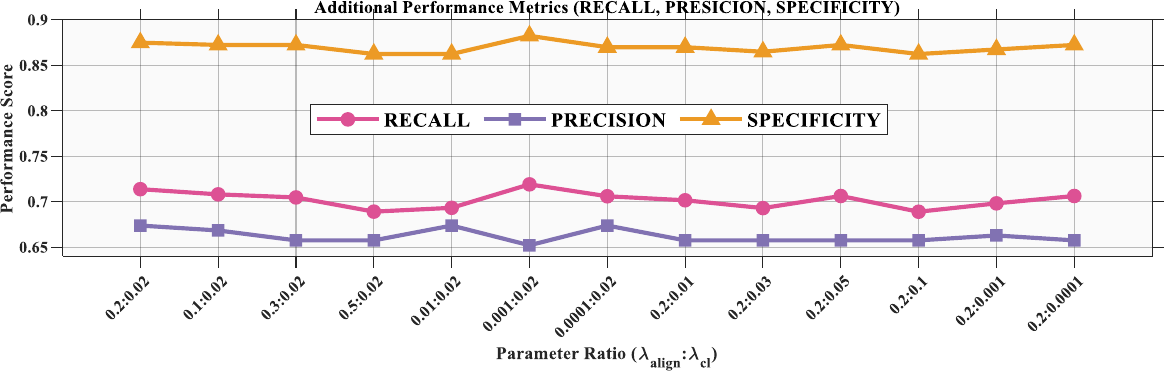}
     \end{subfigure}
     
      \caption{Sensitivity analysis results of loss parameters}
      \label{img3}
     
\end{figure}

\subsubsection{Sensitivity Analysis of the Number of Transformer Encoding Layers and Attention Heads}
To further optimize the structure of the vision-conditioned dual alignment fusion module, we systematically explored the impact of the combination of the number of Transformer layers (1-3 layers) and the number of attention heads (4-32 heads) on model performance. The results are illustrated in Figure \ref{img2}. 
% Figure placeholder
In terms of the number of layers, the 1-layer Transformer architecture showed the best or near-best performance in all head number configurations, with an average AUC of 80.89\%, which is better than the 2-layer (80.04\%) and 3-layer (80.37\%) configurations. This phenomenon indicates that for the image-text cross-modal interaction in this task, a shallow Transformer is sufficient to effectively capture key semantic associations, and increasing the network depth may lead to overfitting due to too many parameters.
In the optimization of the number of attention heads, the 8-head configuration achieved the highest AUC (81.58\%) in the 1-layer architecture, which is better than the 4-head (80.79\%), 16-head (81.06\%), and 32-head (80.11\%) configurations. In the two key indicators of ACC and F1, the 16-head configuration performed best (81.16\%, 69.26\%), and was significantly better than the performance of other head numbers in this layer. This shows that a moderate diversity of attention can most effectively balance the information interaction between different modalities, while too few or too many attention heads will damage the model's performance.
Considering indicators such as AUC, ACC, and F1, we chose a configuration of 1 layer and 16 heads as the final parameters of our model, fully considering the balance between performance and computational power.

\subsubsection{Effect of different losses}
We conducted a systematic grid search in the parameter space ($\lambda_{align}$: $10^{-4}$ to 0.5, $\lambda_{cl}$: $10^{-4}$ to 0.1) to comprehensively evaluate the impact of different parameter combinations on model performance. The sensitivity analysis results in Figure \ref{img3} show that the parameter combination of $\lambda_{align}=0.2$ and $\lambda_{cl}=0.02$ showed the best performance on multiple evaluation indicators. This configuration achieved the highest level on the three key indicators of accuracy (81.16\%), F1-score (69.26\%), and precision (67.39\%). The parameter sensitivity pattern analysis shows that this combination achieved the best balance between the alignment loss and the contrastive learning objective, ensuring the robust performance of the model.

\subsubsection{Sensitivity Analysis of Network Dimension Parameters}
\begin{figure}[!t]
\centering
    \begin{subfigure}{0.49\textwidth}
        \includegraphics[width=1\linewidth]{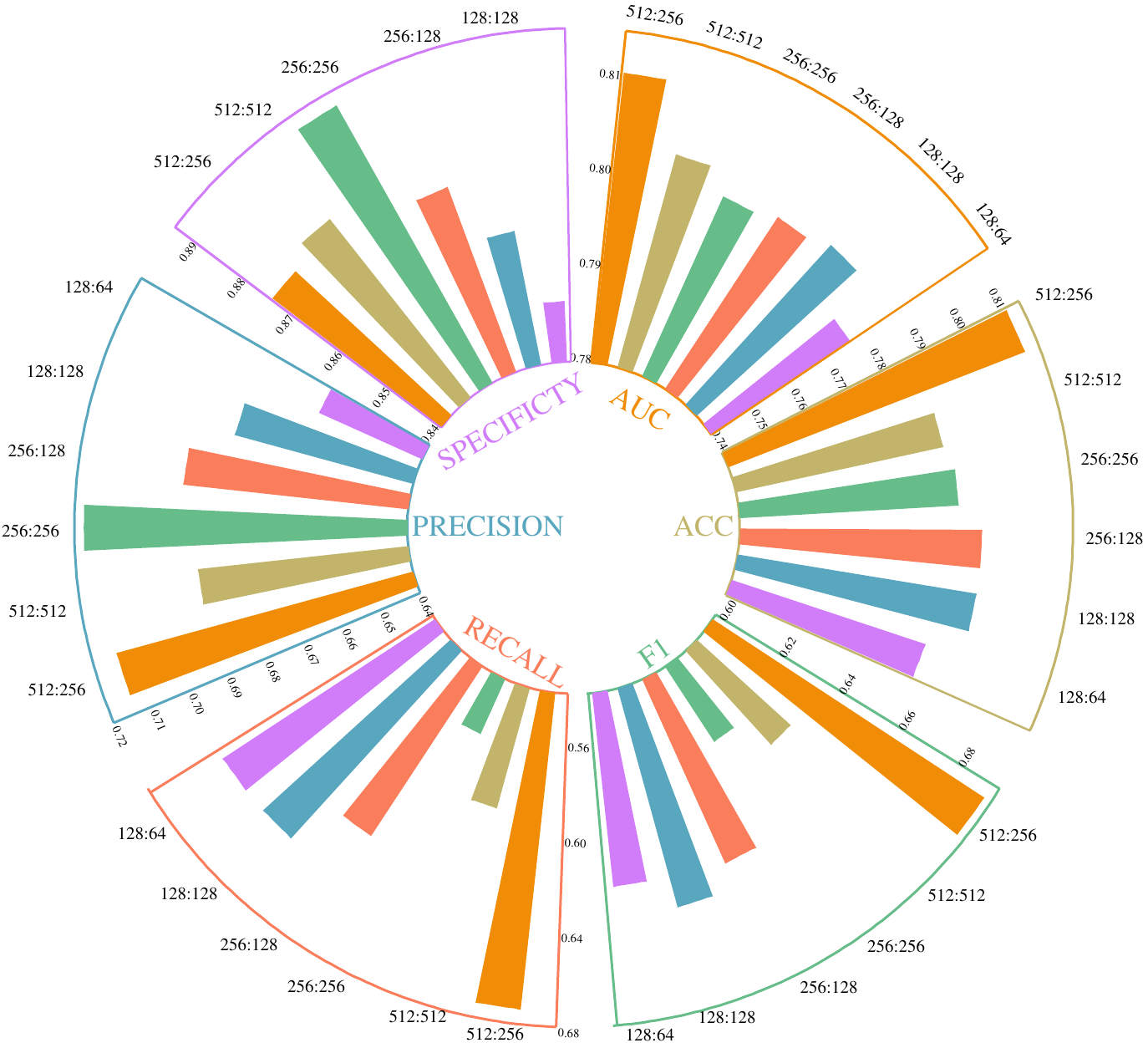}
     \end{subfigure}
     \hspace{0.02\textwidth}
     \begin{subfigure}{0.46\textwidth}
        \includegraphics[width=1\linewidth]{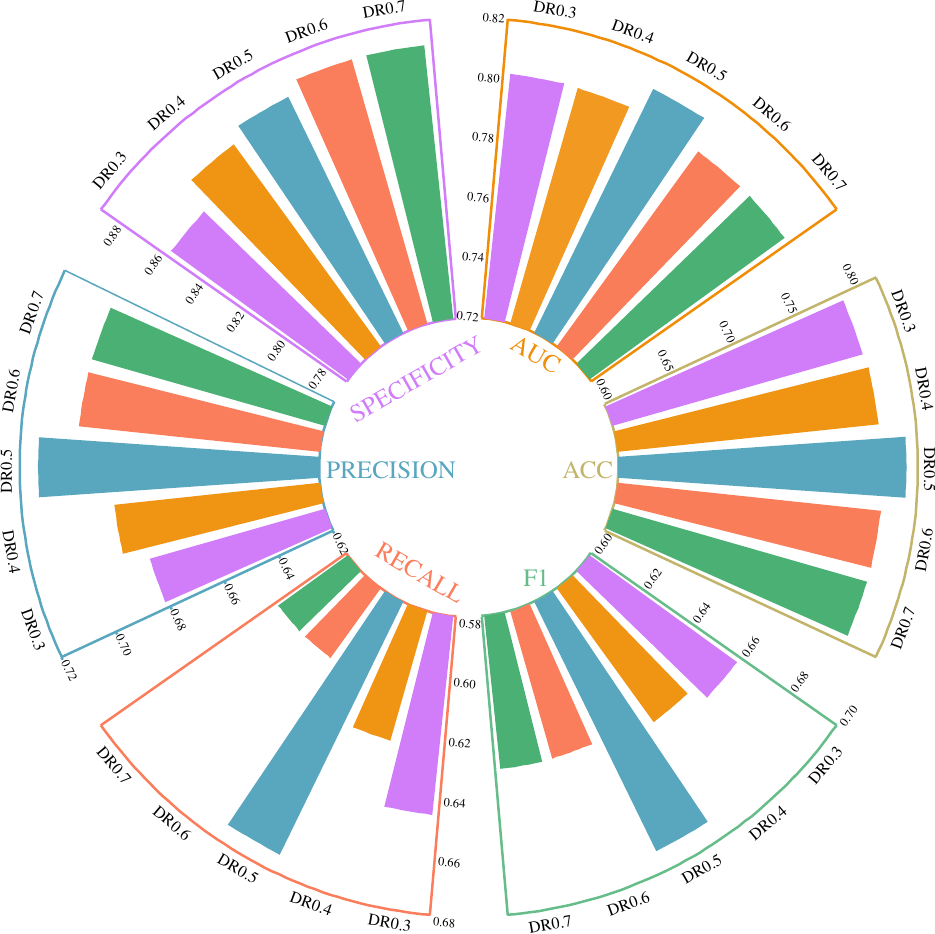}
     \end{subfigure}
     
      \caption{Sensitivity analysis results of network parameters and regularization parameter}
      \label{img4}
     
\end{figure}

For the hidden layer dimension configuration, we conducted a systematic comparison in an expanded architecture space (from 128:64 to 512:512) to evaluate the impact of architectures with different complexities on model performance. As shown in Figure \ref{img4},  The sensitivity analysis results show that the 512:256 architecture ranked first in AUC, accuracy, F1-score, and recall indicators. The architecture sensitivity pattern analysis reveals that higher complexity architectures are prone to overfitting, while lower complexity architectures limit the model's representation ability. The 512:256 architecture achieved a Pareto optimal balance between model complexity and generalization ability.

\subsubsection{Sensitivity Analysis of Regularization Parameters}
The dropout rate was systematically evaluated in the range of (0.3--0.7). The sensitivity analysis clearly shows that a dropout rate of 0.5 achieved the best balance between preventing overfitting and maintaining model capacity. This parameter sensitivity pattern is consistent with the expectations of deep learning regularization theory and was verified in our specific architecture.
Through systematic sensitivity analysis, we not only determined the optimal parameter configuration, but more importantly, revealed the influence patterns of each parameter on model performance, providing a solid basis for the scientific nature of the model design.

\subsection{Robustness and generalizability}

\begin{figure}[!t]
\centering
    \begin{subfigure}{0.45\textwidth}
    \caption{Oriental Hospital}
        \includegraphics[width=1\linewidth]{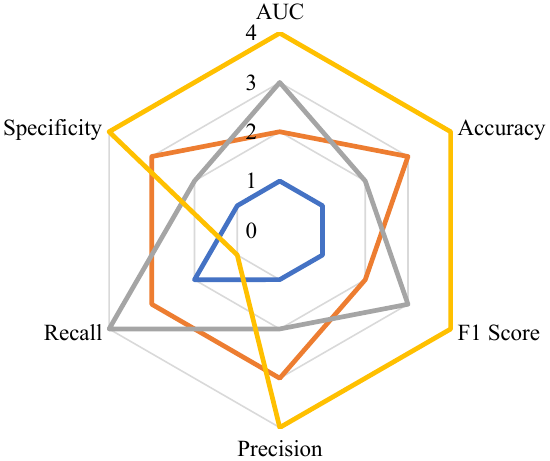}
        
        \label{fig:5a}
     \end{subfigure}
     \hspace{0.02\textwidth}
     \begin{subfigure}{0.45\textwidth}
      \caption{Tongji Hospital}
        \includegraphics[width=1\linewidth]{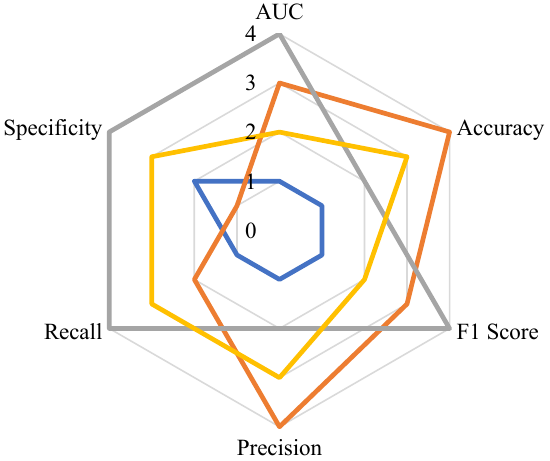}
       
        \label{fig:5b}
     \end{subfigure}
     \begin{subfigure}{0.45\textwidth}
     \caption{Xinhua Hospital}
        \includegraphics[width=1\linewidth]{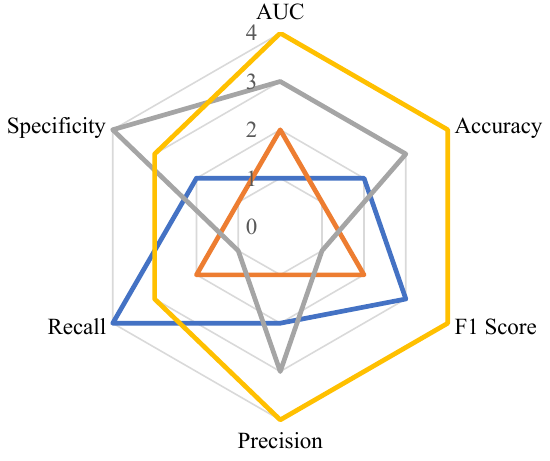}
        
        \label{fig:5c}
     \end{subfigure}
     \begin{subfigure}{0.45\textwidth}
     \caption{Putuo District Central Hospital}
        \includegraphics[width=1\linewidth]{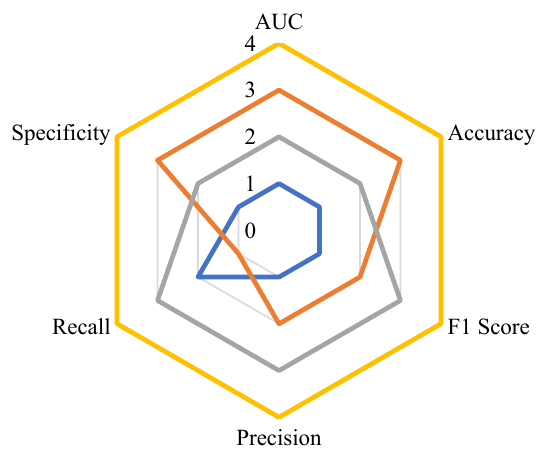}
        
        \label{fig:5d}
     \end{subfigure}

     \vspace{0.25cm}
     
     \begin{subfigure}{0.80\textwidth}
        \includegraphics[width=1\linewidth]{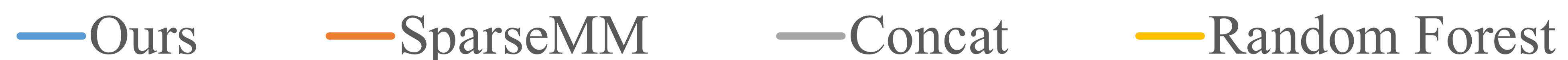}
        \label{fig:5e}
     \end{subfigure}
     \vspace{-0.75cm}
      \caption{Comparative ranking of models under leave-one-hospital-out validation}
      \label{img5}
\end{figure}

To evaluate the clinical translational potential of our proposed framework, we conducted Leave-One-Hospital-Out (LOHO) validation on a multicenter dataset from four institutions (Tongji, Xinhua, Shanghai Oriental, and Putuo Central). As illustrated in Figure \ref{img5}, our model achieved leading AUC and Accuracy across all centers, consistently outperforming SparseMM, Concat, and Random Forest (RF) baselines.

Notably, at Xinhua Hospital (Figure \ref{fig:5c}), our model maintained superior AUC despite lower Precision and Recall. This stems from the extreme class imbalance (131:36) at this center. Unlike traditional Concat or machine learning strategies that are prone to overfitting minority class noise, our core VDAFM module prioritizes global cross-modal semantic consistency through dual alignment. This "robust conservatism" ensures stable feature mapping in scenarios with scarce positive samples, preventing performance degradation by learning a more universal feature distribution.

Overall, the experimental evidence confirms that the VDAFM module effectively captures shared robust features, successfully mitigating Domain Shift caused by device disparities or population variances. Rather than mere local data fitting, our approach achieves a deeper comprehension of intrinsic disease characteristics, demonstrating exceptional generalizability in diverse clinical settings.

\section{Conclusion}
This study addresses key limitations in current ischemic stroke prognosis prediction, which include insufficient multi-modal integration, variable text quality, and inadequate image-text interaction. It innovatively proposes a tri-modal deep learning framework that integrates MRI images, text reports, and clinical tables. The core of this framework is the Vision-Conditioned Dual Alignment Fusion Module (VDAFM), which uses visual representations as conditions and constrains the representations of images and text in both feature and semantic spaces through dual semantic alignment loss and cross-modal contrastive loss, ultimately achieving dynamic and deep fusion of the two modalities. The experimental results demonstrate that fully leveraging large language models to generate high-quality text modalities and integrating tri-modal information with advanced fusion mechanisms can provide a more comprehensive and accurate decision-making basis for stroke prognosis prediction, thereby effectively overcoming the shortcomings of existing methods.

\section{CRediT authorship contribution statement}

 \textbf{Liren Chen:} Conceptualization; Methodology; Software (Code implementation, Core experiments); Validation (Core experiments); Writing – Original Draft (Abstract, Experimental results).
\textbf{Lidong Sun:} Methodology (Traditional experiments); Software/Validation (Model replication); Investigation (Traditional experiments); Writing – Original Draft (Methods section).
\textbf{Mingyan Huang:} Investigation (Literature review); Writing – Original Draft (Introduction, Related Work).
\textbf{Junzhe Tang:} Visualization (Figures optimization); Formal Analysis (Typesetting).
\textbf{Guanjie Wang and Yinghui Zhu:} Investigation (Experiment assistance); Data Curation.
\textbf{Yiqing Xia :} Project Administration (Guidance on submission). 
\textbf{Ting Xiao (Corresponding Author):} Conceptualization (Project direction); Supervision; Writing – Review \& Editing.

\section{Statements and Declarations}

The authors declare that they have no known competing financial interests or personal relationships that could have appeared to influence the work reported in this paper.

\section{Data availability}

Data will be made available on request.

\section{Acknowledgments}

This work is supported by the National Natural Science Foundation of China (No. 62306115).
\bibliographystyle{elsarticle-num} 
 \bibliography{ref}

@article{ref1,
  title={Global, regional, and national burden of stroke and its risk factors, 1990--2021: a systematic analysis for the Global Burden of Disease Study 2021},
  author={Feigin, Valery L and Abate, Melsew Dagne and Abate, Yohannes Habtegiorgis and Abd ElHafeez, Samar and Abd-Allah, Foad and Abdelalim, Ahmed and Abdelkader, Atef and Abdelmasseh, Michael and Abd-Elsalam, Sherief and Abdi, Parsa and others},
  journal={The Lancet Neurology},
  volume={23},
  number={10},
  pages={973--1003},
  year={2024},
  publisher={Elsevier}
}

@article{ref2,
  title={Global, regional, and national trends in ischaemic stroke burden and risk factors among adults aged 20+ years (1990--2021): a systematic analysis of data from the Global Burden of Disease study 2021 with projections into 2050},
  author={Liu, Sibo and Li, Yanzhao and Lan, Xiaoyan and Wang, Long and Li, Hang and Gu, Dean and Wang, Mengxing and Liu, Jinjie},
  journal={Frontiers in Public Health},
  volume={13},
  pages={1567275},
  year={2025},
  publisher={Frontiers Media SA}
}

@article{ref3,
  title={Prognosis of ischemic stroke predicted by machine learning based on multi-modal MRI radiomics},
  author={Yu, Huan and Wang, Zhenwei and Sun, Yiqing and Bo, Wenwei and Duan, Kai and Song, Chunhua and Hu, Yi and Zhou, Jie and Mu, Zizhang and Wu, Ning},
  journal={Frontiers in Psychiatry},
  volume={13},
  pages={1105496},
  year={2023},
  publisher={Frontiers Media SA}
}

@article{ref4,
  title={Functional outcome prediction in ischemic stroke: a comparison of machine learning algorithms and regression models},
  author={Alaka, Shakiru A and Menon, Bijoy K and Brobbey, Anita and Williamson, Tyler and Goyal, Mayank and Demchuk, Andrew M and Hill, Michael D and Sajobi, Tolulope T},
  journal={Frontiers in neurology},
  volume={11},
  pages={889},
  year={2020},
  publisher={Frontiers Media SA}
}

@inproceedings{ref5,
  title={Random decision forests},
  author={Ho, Tin Kam},
  booktitle={Proceedings of 3rd international conference on document analysis and recognition},
  volume={1},
  pages={278--282},
  year={1995},
  organization={IEEE}
}

@article{ref6,
  title={Decision tree-based diagnosis of coronary artery disease: CART model},
  author={Ghiasi, Mohammad M and Zendehboudi, Sohrab and Mohsenipour, Ali Asghar},
  journal={Computer methods and programs in biomedicine},
  volume={192},
  pages={105400},
  year={2020},
  publisher={Elsevier}
}

@article{ref7,
  title={CMBA-SVM: a clinical approach for Parkinson disease diagnosis},
  author={Sahu, Bibhuprasad and Mohanty, Sachi Nandan},
  journal={International Journal of Information Technology},
  volume={13},
  number={2},
  pages={647--655},
  year={2021},
  publisher={Springer}
}

@article{ref8,
  title={Logistic regression was as good as machine learning for predicting major chronic diseases},
  author={Nusinovici, Simon and Tham, Yih Chung and Yan, Marco Yu Chak and Ting, Daniel Shu Wei and Li, Jialiang and Sabanayagam, Charumathi and Wong, Tien Yin and Cheng, Ching-Yu},
  journal={Journal of clinical epidemiology},
  volume={122},
  pages={56--69},
  year={2020},
  publisher={Elsevier}
}

@article{add1,
  title={Multimodal machine learning for stroke prognosis and diagnosis: A systematic review},
  author={Shurrab, Saeed and Guerra-Manzanares, Alejandro and Magid, Amani and Piechowski-Jozwiak, Bartlomiej and Atashzar, S Farokh and Shamout, Farah E},
  journal={IEEE Journal of Biomedical and Health Informatics},
  year={2024},
  publisher={IEEE}
}

@article{add2,
  title={Prognosis of ischemic stroke predicted by machine learning based on multi-modal MRI radiomics},
  author={Yu, Huan and Wang, Zhenwei and Sun, Yiqing and Bo, Wenwei and Duan, Kai and Song, Chunhua and Hu, Yi and Zhou, Jie and Mu, Zizhang and Wu, Ning},
  journal={Frontiers in Psychiatry},
  volume={13},
  pages={1105496},
  year={2023},
  publisher={Frontiers Media SA}
}

@inproceedings{ref9,
  title={Transformer-based classification outcome prediction for multimodal stroke treatment},
  author={Ma, Danqing and Wang, Meng and Xiang, Ao and Qi, Zongqing and Yang, Qin},
  booktitle={2024 IEEE 2nd International Conference on Sensors, Electronics and Computer Engineering (ICSECE)},
  pages={383--386},
  year={2024},
  organization={IEEE}
}

@article{ref10,
  title={Multimodal predictive modeling of endovascular treatment outcome for acute ischemic stroke using machine-learning},
  author={Brugnara, Gianluca and Neuberger, Ulf and Mahmutoglu, Mustafa A and Foltyn, Martha and Herweh, Christian and Nagel, Simon and Sch{\"o}nenberger, Silvia and Heiland, Sabine and Ulfert, Christian and Ringleb, Peter Arthur and others},
  journal={Stroke},
  year={2020},
  publisher={Lippincott Williams \& WilkinsHagerstown, MD}
}

@inproceedings{ref11,
  title={Prediction of thrombectomy functional outcomes using multimodal data},
  author={Samak, Zeynel A and Clatworthy, Philip and Mirmehdi, Majid},
  booktitle={Annual Conference on Medical Image Understanding and Analysis},
  pages={267--279},
  year={2020},
  organization={Springer}
}

@article{ref12,
  title={Precision medicine in stroke: towards personalized outcome predictions using artificial intelligence},
  author={Bonkhoff, Anna K and Grefkes, Christian},
  journal={Brain},
  volume={145},
  number={2},
  pages={457--475},
  year={2022},
  publisher={Oxford University Press}
}

@misc{ref13,
  title={Fusion: Fully integration of vision-language representations for deep cross-modal understanding},
  author={Liu, Zheng and Liu, Mengjie and Chen, Jingzhou and Xu, Jingwei and Cui, Bin and He, Conghui and Zhang, Wentao},
  journal={arXiv preprint arXiv:2504.09925},
  year={2025}
}

@article{ref14,
  title={Multi-view multi-scale CNNs for lung nodule type classification from CT images},
  author={Liu, Xinglong and Hou, Fei and Qin, Hong and Hao, Aimin},
  journal={Pattern Recognition},
  volume={77},
  pages={262--275},
  year={2018},
  publisher={Elsevier}
}

@article{add4,
  title={Natural language processing in medicine: a review},
  author={Locke, Saskia and Bashall, Anthony and Al-Adely, Sarah and Moore, John and Wilson, Anthony and Kitchen, Gareth B},
  journal={Trends in Anaesthesia and Critical Care},
  volume={38},
  pages={4--9},
  year={2021},
  publisher={Elsevier}
}

@inproceedings{add5,
  title={Multi-Modal Large Language Models are Effective Vision Learners},
  author={Sun, Li and Ahuja, Chaitanya and Chen, Peng and D'Zmura, Matt and Batmanghelich, Kayhan and Bontrager, Philip},
  booktitle={2025 IEEE/CVF Winter Conference on Applications of Computer Vision (WACV)},
  pages={8617--8626},
  year={2025},
  organization={IEEE}
}

@article{add6,
  title={Csan: cross-coupled semantic adversarial network for cross-modal retrieval},
  author={Li, Zhuoyi and Lu, Huibin and Fu, Hao and Meng, Fanzhen and Gu, Guanghua},
  journal={Artificial Intelligence Review},
  volume={58},
  number={5},
  pages={1--27},
  year={2025},
  publisher={Springer}
}

@inproceedings{ref15,
  title={Classification of breast MRI lesions using small-size training sets: comparison of deep learning approaches},
  author={Amit, Guy and Ben-Ari, Rami and Hadad, Omer and Monovich, Einat and Granot, Noa and Hashoul, Sharbell},
  booktitle={Medical Imaging 2017: Computer-Aided Diagnosis},
  volume={10134},
  pages={374--379},
  year={2017},
  organization={SPIE}
}

@article{ref16,
  title={Dermatologist-level classification of skin cancer with deep neural networks},
  author={Esteva, Andre and Kuprel, Brett and Novoa, Roberto A and Ko, Justin and Swetter, Susan M and Blau, Helen M and Thrun, Sebastian},
  journal={nature},
  volume={542},
  number={7639},
  pages={115--118},
  year={2017},
  publisher={Nature Publishing Group UK London}
}

@article{ref17,
  title={Prediction of tissue outcome and assessment of treatment effect in acute ischemic stroke using deep learning},
  author={Nielsen, Anne and Hansen, Mikkel Bo and Tietze, Anna and Mouridsen, Kim},
  journal={Stroke},
  volume={49},
  number={6},
  pages={1394--1401},
  year={2018},
  publisher={Lippincott Williams \& Wilkins Hagerstown, MD}
}

@article{ref18,
  title={Predicting 6-month unfavorable outcome of acute ischemic stroke using machine learning},
  author={Li, Xiang and Pan, XiDing and Jiang, ChunLian and Wu, MingRu and Liu, YuKai and Wang, FuSang and Zheng, XiaoHan and Yang, Jie and Sun, Chao and Zhu, YuBing and others},
  journal={Frontiers in neurology},
  volume={11},
  pages={539509},
  year={2020},
  publisher={Frontiers Media SA}
}

@article{ref19,
  title={Machine learning--based model for prediction of outcomes in acute stroke},
  author={Heo, JoonNyung and Yoon, Jihoon G and Park, Hyungjong and Kim, Young Dae and Nam, Hyo Suk and Heo, Ji Hoe},
  journal={Stroke},
  volume={50},
  number={5},
  pages={1263--1265},
  year={2019},
  publisher={Lippincott Williams \& Wilkins Hagerstown, MD}
}

@article{ref20,
  title={Long-term cancer survival prediction using multimodal deep learning},
  author={Vale-Silva, Lu{\'\i}s A and Rohr, Karl},
  journal={Scientific Reports},
  volume={11},
  number={1},
  pages={13505},
  year={2021},
  publisher={Nature Publishing Group UK London}
}

@article{ref21,
  title={Predicting Alzheimer’s disease progression using multi-modal deep learning approach},
  author={Lee, Garam and Nho, Kwangsik and Kang, Byungkon and Sohn, Kyung-Ah and Kim, Dokyoon},
  journal={Scientific reports},
  volume={9},
  number={1},
  pages={1952},
  year={2019},
  publisher={Nature Publishing Group UK London}
}

@article{ref22,
  title={A system for accurately predicting the risk of myocardial infarction using PCG, ECG and clinical features},
  author={Zarrabi, Maryam and Parsaei, Hossien and Boostani, Reza and Zare, Azam and Dorfeshan, Zhila and Zarrabi, Khalil and Kojuri, Javad},
  journal={Biomedical Engineering: Applications, Basis and Communications},
  volume={29},
  number={03},
  pages={1750023},
  year={2017},
  publisher={World Scientific}
}

@inproceedings{ref23,
  title={Stroke Outcome Prediction via Multi-level Feature and Multi-modal Fusion Network},
  author={Xiao, Ting and Shi, Lei and Wang, Hao and Wang, Zhe and Lin, Yi},
  booktitle={2024 IEEE International Conference on Bioinformatics and Biomedicine (BIBM)},
  pages={6732--6739},
  year={2024},
  organization={IEEE}
}

@article{ref24,
  title={Automatic prediction of stroke treatment outcomes: latest advances and perspectives},
  author={Samak, Zeynel A and Clatworthy, Philip and Mirmehdi, Majid},
  journal={Biomedical Engineering Letters},
  pages={1--22},
  year={2025},
  publisher={Springer}
}

@article{ref25,
  title={SparseMM: Head Sparsity Emerges from Visual Concept Responses in MLLMs},
  author={Wang, Jiahui and Liu, Zuyan and Rao, Yongming and Lu, Jiwen},
  journal={arXiv preprint arXiv:2506.05344},
  year={2025}
}

@article{ref26,
  title={EarthMind: Towards Multi-Granular and Multi-Sensor Earth Observation with Large Multimodal Models},
  author={Shu, Yan and Ren, Bin and Xiong, Zhitong and Paudel, Danda Pani and Van Gool, Luc and Demir, Begum and Sebe, Nicu and Rota, Paolo},
  journal={arXiv preprint arXiv:2506.01667},
  year={2025}
}

@article{ref27,
  title={Collaborative multi-modal deep learning and radiomic features for classification of strokes within 6 h},
  author={Yoon, Chiho and Misra, Sampa and Kim, Kwang-Ju and Kim, Chulhong and Kim, Bum Joon},
  journal={Expert Systems with Applications},
  volume={228},
  pages={120473},
  year={2023},
  publisher={Elsevier}
}

@article{add7,
  title={C 2 MA-Net: Cross-modal cross-attention network for acute ischemic stroke lesion segmentation based on CT perfusion scans},
  author={Shi, Tianyu and Jiang, Huiyan and Zheng, Bin},
  journal={IEEE Transactions on Biomedical Engineering},
  volume={69},
  number={1},
  pages={108--118},
  year={2021},
  publisher={IEEE}
}

@article{add8,
  title={A study of CNN and transfer learning in medical imaging: Advantages, challenges, future scope},
  author={Salehi, Ahmad Waleed and Khan, Shakir and Gupta, Gaurav and Alabduallah, Bayan Ibrahimm and Almjally, Abrar and Alsolai, Hadeel and Siddiqui, Tamanna and Mellit, Adel},
  journal={Sustainability},
  volume={15},
  number={7},
  pages={5930},
  year={2023},
  publisher={MDPI}
}

@article{ref28,
  title={Toward safer highways, application of XGBoost and SHAP for real-time accident detection and feature analysis},
  author={Parsa, Amir Bahador and Movahedi, Ali and Taghipour, Homa and Derrible, Sybil and Mohammadian, Abolfazl Kouros},
  journal={Accident Analysis \& Prevention},
  volume={136},
  pages={105405},
  year={2020},
  publisher={Elsevier}
}

@article{ref29,
  title={Learning representations by back-propagating errors},
  author={Rumelhart, David E and Hinton, Geoffrey E and Williams, Ronald J},
  journal={nature},
  volume={323},
  number={6088},
  pages={533--536},
  year={1986},
  publisher={Nature Publishing Group UK London}
}

\end{document}